\documentclass[runningheads]{llncs}

\usepackage{eccv}

\usepackage{eccvabbrv}

\usepackage{graphicx}
\usepackage{booktabs}
\usepackage{microtype}
\usepackage{multirow}
\usepackage[numbers,sort&compress]{natbib}
\usepackage{xcolor} 
\usepackage{graphicx} 
\usepackage{placeins}
\newcommand{\res}[1]{{\color{darkgray}\scalebox{0.7}{$\pm$#1}}}

\usepackage[accsupp]{axessibility}  

\usepackage{hyperref}

\usepackage{orcidlink}

\begin{document}

\newcommand{\modelname}{Pathryoshka}
\title{Pathryoshka: Compressing Pathology Foundation Models via Multi-Teacher Knowledge Distillation with Nested Embeddings} 

\titlerunning{Pathryoshka: Multi-Teacher Distillation with Nested Embeddings}

\author{Christian Grashei\inst{1,2,3}\fnmsep\thanks{C.~Grashei and C.~Brechenmacher contributed equally.}\orcidlink{0009-0005-4578-8465} \and
Christian Brechenmacher\inst{4}\fnmsep\protect\footnotemark[1]\orcidlink{0009-0005-8415-6735} \and
Rao Muhammad Umer\inst{4} \and
Jingsong Liu\inst{1,3}\orcidlink{0009-0002-3174-3352} \and
Carsten Marr\inst{4,6,7,8}\fnmsep\thanks{C.~Marr, P.~Schüffler and E.~Szczurek supervised equally.}\orcidlink{0000-0003-2154-4552} \and
Peter Schüffler\inst{1,2,3,8}\fnmsep\protect\footnotemark[2]\orcidlink{0000-0002-1353-8921} \and
Ewa Szczurek\inst{4,5}\fnmsep\protect\footnotemark[2]\orcidlink{0000-0002-1320-6695}}

\authorrunning{C.~Grashei et al.}

\institute{Institute of Pathology, Technical University of Munich, Munich, Germany
\and
Munich Data Science Institute, Munich, Germany
\and
Munich Center for Machine Learning, Munich, Germany
\and
Institute of AI for Health, Helmholtz Munich, Munich, Germany
\and
Institute of Informatics, University of Warsaw, Warsaw, Poland
\and
Dep. of Medicine III, Ludwig-Maximilian-University Hospital, Munich, Germany
\and
Department of Physics, Ludwig-Maximilian-University, Munich, Germany
\and
German Cancer Consortium (DKTK), Munich, Germany
}

\maketitle

\begingroup
\renewcommand\thefootnote{}
\footnotetext{19th European Conference on Computer Vision (ECCV 2026).}
\addtocounter{footnote}{-1}
\endgroup

\begin{abstract}
Foundation models (FMs) have driven significant progress in computational pathology. These models can easily exceed a billion parameters and produce high-dimensional embeddings, thus limiting their applicability for research or clinical use when computing resources are tight. Here we introduce Pathryoshka, a novel multi-teacher distillation framework inspired by agglomerative models and Matryoshka representation learning to reduce pathology FM sizes while allowing for adaptable embedding dimensions. We evaluate our framework with a distilled model on ten public pathology benchmarks with varying downstream tasks. Compared to its much larger teachers, Pathryoshka reduces the model size by 86-92\% at on-par performance. It outperforms state-of-the-art single-teacher distillation models of comparable size by a median margin of 7.0 and other pathology multi-teacher distillation by 5.3 percentage points in accuracy. By enabling efficient deployment without sacrificing accuracy or representational richness, Pathryoshka democratizes access to state-of-the-art pathology FMs for the broader research and clinical community.
  
\end{abstract}

\section{Introduction}
\label{sec:intro}

\begin{figure}[tb]
    \centering
    \begin{subfigure}[c]{0.45\textwidth}
        \centering
        \includegraphics[width=\linewidth]{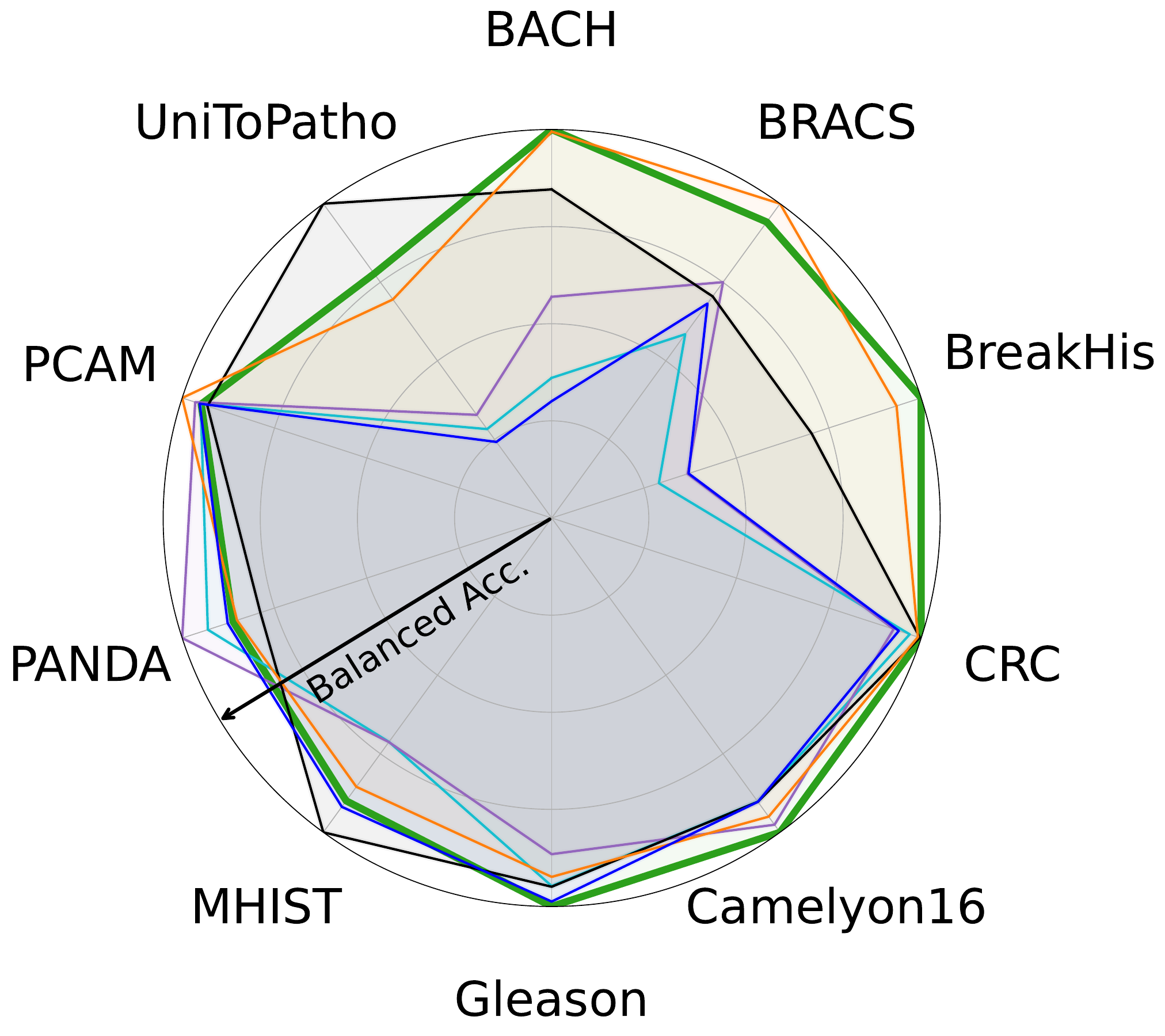}
    \end{subfigure}
    \begin{subfigure}[c]{0.50\textwidth}
        \centering
        \begin{tikzpicture}
        \node[
          draw=black,
          line width=.2pt,
          rounded corners=1mm,
          inner sep=1mm  
        ] {
            \includegraphics[width=\linewidth]{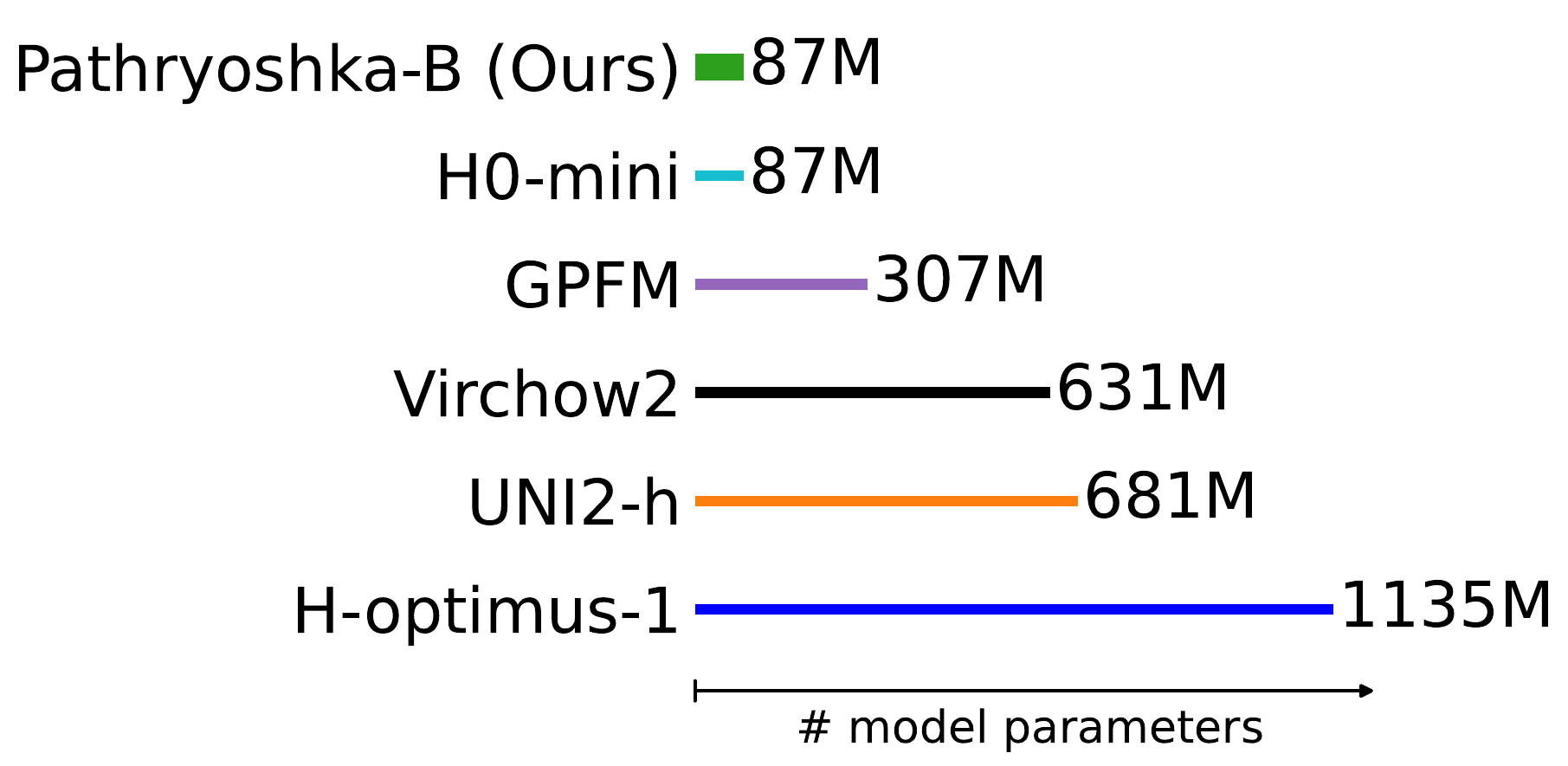}
        };
        \end{tikzpicture}
    \end{subfigure}
    \caption{
        Although smaller in parameter size, \modelname{} outperforms or is on par with its state-of-the-art teacher pathology FMs on various benchmarks and outperforms current single-teacher (H0-mini) and multi-teacher-distillation (GPFM) models.
    }
    \label{fig:radar_benchmarks}
\end{figure}

Empirical scaling laws demonstrate consistent performance gains from increasing foundation model (FM) size.
Larger models capture more complex structures and show stronger performance across tasks~\cite{villalobos2022machinelearningmodelsizes, pmlr-v139-hashimoto21a,kaplan2020scalinglawsneurallanguage}.
Yet, model scaling also brings rapid cost escalation, resulting in inaccessible state-of-the-art development and deployment. 
This challenge is acute in pathology: Although pathology FMs promise major advances in cancer diagnosis and patient care \cite{bilal2025foundationmodelscomputationalpathology,ochi2024pathologyfoundationmodels}, their hardware and operational costs hinder  adoption in  resource-limited clinics and research centers.  Pathology FMs contain hundreds of millions to billions of parameters, and with whole‑slide images requiring thousands of tiles to be encoded, inference and storage costs become prohibitive. Thus, there is a clear need for pathology FMs with reduced model and embedding sizes that maintain performance while enabling compute- and memory-efficient deployment.

While several promising approaches exist in this direction, they still face notable limitations.
Naively training smaller models on equally rich datasets does not match the performance of larger models~\cite{10.1007/978-3-031-44917-8_25}. A prominent approach to reducing  model size while preserving accuracy is \textit{knowledge distillation}, which transfers knowledge from large teacher models   to a smaller student  by aligning their embeddings or logits~\cite{Hinton2015-io,ranzinger2024radio,heinrich2025radiov2,ma2025generalizable}.
Although highly effective compared to naive approaches such as model ensembling, and able to substantially reduce model size~\cite{Neidlinger2024-of,chen2024uni}, distillation typically retains large embedding sizes, leading to high inference times and memory usage in downstream tasks. 
To obtain compact embeddings, {\textit{Matryoshka representation learning} (MRL)~\citep{kusupati2022matryoshka,venkataramanan2025franca} trains a single high-dimensional embedding whose prefix sub-vectors remain semantically meaningful and effective. This nested structure enables  dynamic   truncation to lower dimensions with minimal loss. However, while MRL  compresses embeddings, it does not guarantee overall model size reduction.  To our knowledge, no prior work has combined distillation with MRL to jointly realize the benefits of both approaches.  

In the context of pathology FMs,  distillation is often performed using a single teacher model, such as H0-mini~\cite{h0-mini} trained by distillation from a large vision transformer model called H-optimus-0~\cite{hoptimus0}. However,  relying on a single teacher risks inheriting its biases and limits the represented diversity of clinical variation. We illustrate this behavior in  \cref{fig:radar_benchmarks}, where the distilled H0-mini is only able to perform as good or only marginally better than H-optimus-1~\cite{hoptimus1}, a model trained as successor to H-optimus-0. While multi-teacher distillation strategies such as AM-RADIO~\cite{ranzinger2024radio} have been successfully applied to general image foundation models, they are underexplored in pathology. Combining multiple pathology FMs has been done with GPFM~\cite{ma2025generalizable}, but without reduction in model capacity with respect to the teacher models. In summary, pathology FMs still lack a method that simultaneously compresses both model and embedding size while leveraging diverse multi-teacher signals.

To address these challenges, we propose \modelname{}, a pathology FM that leverages multi-teacher knowledge distillation in combination with MRL to fundamentally reduce model size and improve computational efficiency while achieving similar performances to large teacher models~\cite{chen2024uni,zimmermann2024virchow2, hoptimus1, conch} and outperforming previous distilled small models~\cite{h0-mini} (\cref{fig:radar_benchmarks}).
The hierarchical embedding structure of \modelname{} fuses complementary signals from multiple teachers into compact and robust subembeddings. To avoid memorization of embeddings of teacher models from public training data while ensuring enough data diversity, distillation training of \modelname{} is performed on an internal dataset of 243 million image tiles extracted from 158,233 H\&E-stained images across all organs, which is augmented using a novel cropping strategy for distillation. 
Evaluations on multiple benchmarks, including patch-level and slide-level, demonstrate that \modelname{} maintains parity or outperforms individual teacher models in terms of efficiency and performance. Code and model weights are available at \url{https://huggingface.co/SchuefflerLab/Pathryoshka-B}.

Our main contributions are the following:

\begin{itemize}
    \item We introduce a multi-teacher distillation framework for unsupervised distillation of pathology FMs that relies solely on CLS and patch tokens, enabling the use of external public models. 
    \item We propose nested embeddings for multi-teacher distillation, yielding a pathology FM with adaptable embeddings.
    \item We demonstrate  that multiple large-scale pathology FM teacher models can be effectively compressed into a single compact model with competitive performance, substantially reducing model size and inference cost.
    \item We introduce a novel cropping-based augmentation strategy for multi-teacher distillation improving representation alignment across teachers. 
\end{itemize}

\section{Related Work}

\subsubsection{Current Pathology Foundation Models.}
Recent years have witnessed a surge of pathology FMs leveraging vision transformer architecture~\cite{dosovitskiy2020image} and self-supervised learning on large unlabeled datasets to learn generalizable representations that can be efficiently adapted to many downstream pathology tasks with minimal labeled data~\cite{wangTransformerbasedUnsupervisedContrastive2022,filiotScalingSelfSupervisedLearning2023,filiotPhikonv2LargePublic2024,vorontsov2024foundation,zimmermann2024virchow2,chen2024uni,hoptimus0,luVisuallanguageFoundationModel2024}. Among most prominent pathology FMs, three large models have shown excellent performance: Virchow2, UNI2-h, and H-optimus-1. Virchow2~\cite{zimmermann2024virchow2} is a 632-million-parameter vision transformer model (ViT-H/14) for pathology, extending the earlier Virchow model~\cite{vorontsov2024foundation} and pretrained on an exceptionally large dataset of 3.1 million H\&E-stained whole-slide images (WSIs) from 225,000 patients, totaling 2 billion image tiles. A bigger variant, Virchow2G (1.9 billion parameters, ViT-G), was introduced to examine pure model scaling effects. UNI2-h~\cite{chen2024uni} is a 681-million-parameter model (ViT-H/14) pretrained on over 200 million tiles from more than 350,000 H\&E WSIs across 20 major tissue types. H-optimus-1~\cite{hoptimus1} is a 1.1-billion-parameter ViT-G/14 model pretrained on over one million H\&E WSIs, comprising billions of tiles. Despite these advances, adoption of current resource-hungry pathology FMs in the clinic remains limited, and no single model provides both optimal efficiency and performance across all pathology tasks.

\subsubsection{Distillation for Pathology Foundation Models.}
In contrast to the extensive literature on supervised distillation, unsupervised knowledge distillation represents an underexplored frontier. This research gap extends to the pathology domain, which has seen a surge of FMs trained in an unsupervised manner. Given the scarcity of annotated data, the importance of label-free distillation methods is amplified, as they provide a vital mechanism for compressing foundational knowledge without the need for supervision labels.

So far, only \textit{single}-teacher distillation has been studied for reducing pathology FM size. One example is H0-mini \cite{h0-mini}, a model distilled from H-optimus-0 (ViT-G) into a smaller (ViT-B) architecture with 86M parameters. Virchow2G-Mini (ViT-S), using 22M parameters, is distilled from Virchow2G (ViT-G). Notably, both models rely on the DINOv2 distillation framework~\cite{oquab2023dinov2}, which necessitates access to teacher heads and prototype scores that are frequently omitted from public model releases, limiting the reproducibility and flexibility of the distillation process. Single-teacher distillation additionally inherits the weaknesses of its teacher: Figure \ref{fig:radar_benchmarks} demonstrates that H0-mini performs similarly to H-optimus-1 from the same model family. GPFM \cite{ma2025generalizable} introduces a multi-teacher objective while also integrating DINO \cite{Caron_2021_ICCV} and Masked Image Modeling \cite{he2022masked} losses. Its student architecture is of same size or larger in scale to its teachers, thus not addressing model efficiency or significant parameter compression.

Consequently, the distillation of knowledge from multiple, substantially larger teachers into a compact student remains an open research question. Furthermore, we posit that unsupervised distillation can serve as more than a compression tool. It can act as a knowledge agglomeration to create pathology FMs that surpass the performance of individual teachers. Finally, the integration of nested embeddings within a distillation framework remains unexplored in the pathology domain, representing a missed opportunity for creating flexible representations. Flexible embeddings offer a trade-off between accuracy and efficiency where resources are scarce. Working with gigapixel whole-slide images often involves storing large amounts of embeddings in memory or disk~\cite{chen2022fast, schirris2022deepsmile}. Reducing the embedding size alleviates the storage burden while accelerating downstream tasks where compute scales with dimensionality, such as clustering for large-scale data curation \cite{vo2024automatic}.

\section{Methodology}
\label{sec:method}

\begin{figure}[tb]
  \centering
   \includegraphics[width=0.95\linewidth]{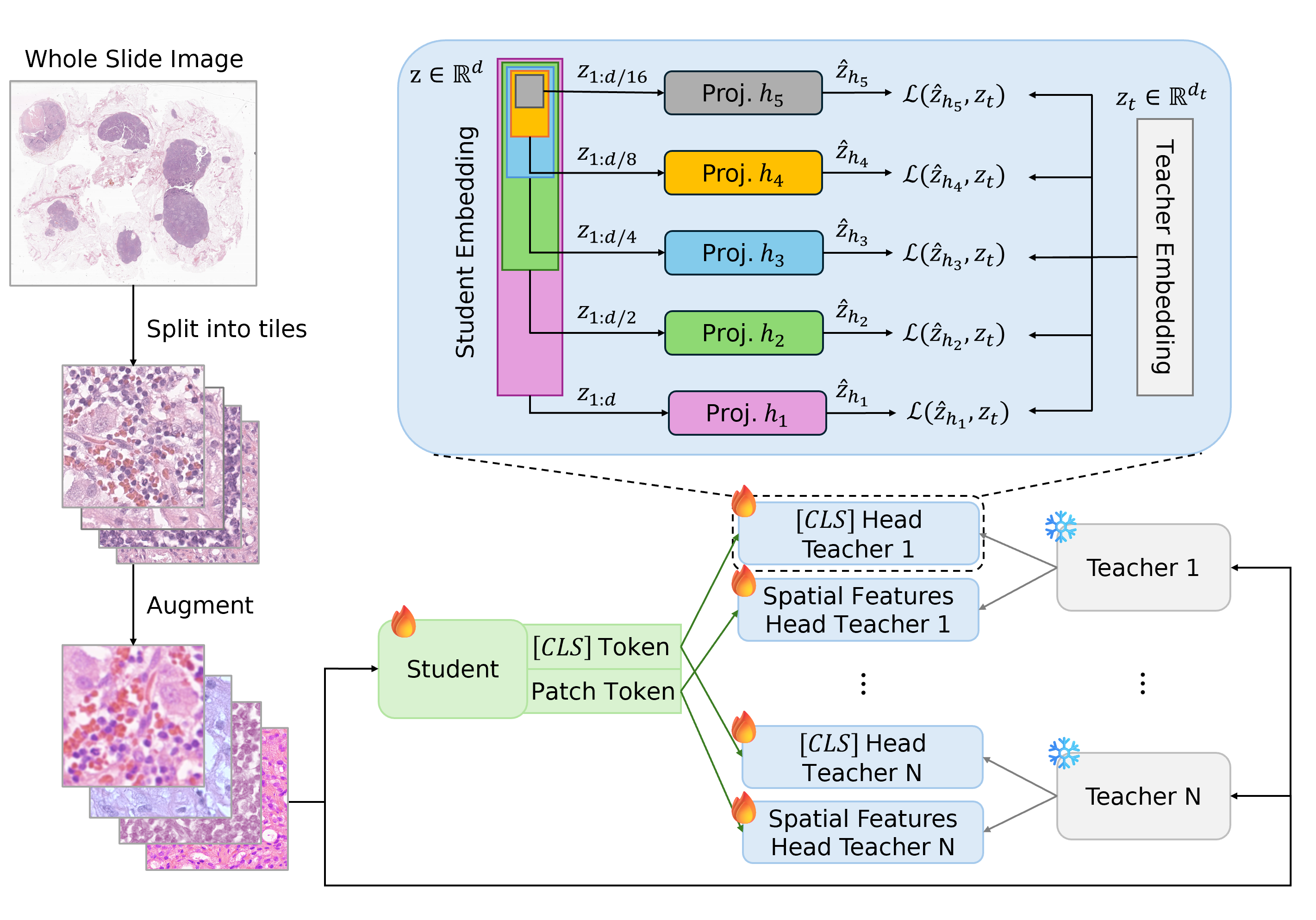}
   \caption{In \modelname, a multi-teacher distillation framework, the student model and distillation heads are jointly trained while teacher models remain frozen. Input images are augmented and processed by both the student and all teachers. The CLS and spatial feature heads align student representations with those of the teachers. Each head includes multiple MLP projection layers that form nested embeddings by minimizing a similarity loss between student and teacher projections.}
   \label{fig:distillation_overview}
\end{figure}

We introduce \modelname, a framework for distilling multiple teacher FMs into a single, compact vision transformer (Fig.~\ref{fig:distillation_overview}). The only prerequisite for the teachers is the provision of token-based representations, specifically a global class (CLS) token  and spatial patch tokens, to align with the student's output.
Pathryoshka draws inspiration from the AM-RADIO~\cite{ranzinger2024radio} approach for agglomerative distillation using multiple heterogeneous teachers, as well as from Matryoshka representation learning~\cite{kusupati2022matryoshka} for flexible embeddings of varying dimensionalities. While demonstrated for histopathology, our proposed setup is domain-agnostic, enabling application in other fields where multiple homogeneous models are available. It operates in a fully unsupervised manner, requiring only a large-scale unlabeled dataset.

\subsection{Dataset}
To train our model, we use a large-scale in-house dataset of 158,233 H\&E stained whole-slide images (WSIs). We identify tissue regions using CLAM~\cite{lu2021data}. We randomly sample a maximum of 2,250 tiles of size 448x448 pixels  per WSI at 10x, 20x and 40x magnifications in proportions of 20\%, 40\% and 40\%, respectively, to cover multiple morphological scales. If there was not enough tissue on a WSI, sampling is stopped earlier, resulting in a total of 243 million tiles. While substantial in absolute terms, large-scale foundation models often exceed this training dataset size~\cite{hoptimus1,zimmermann2024virchow2}. We intentionally exclude public datasets from training to ensure a fair and unbiased evaluation. In the context of distillation, separating training and evaluation data is particularly important to avoid the student model to memorize teacher embeddings for seen data, and to be able to demonstrate generalization to unseen inputs.

\begin{figure}[tb]
  \centering
   \includegraphics[width=0.55\linewidth]{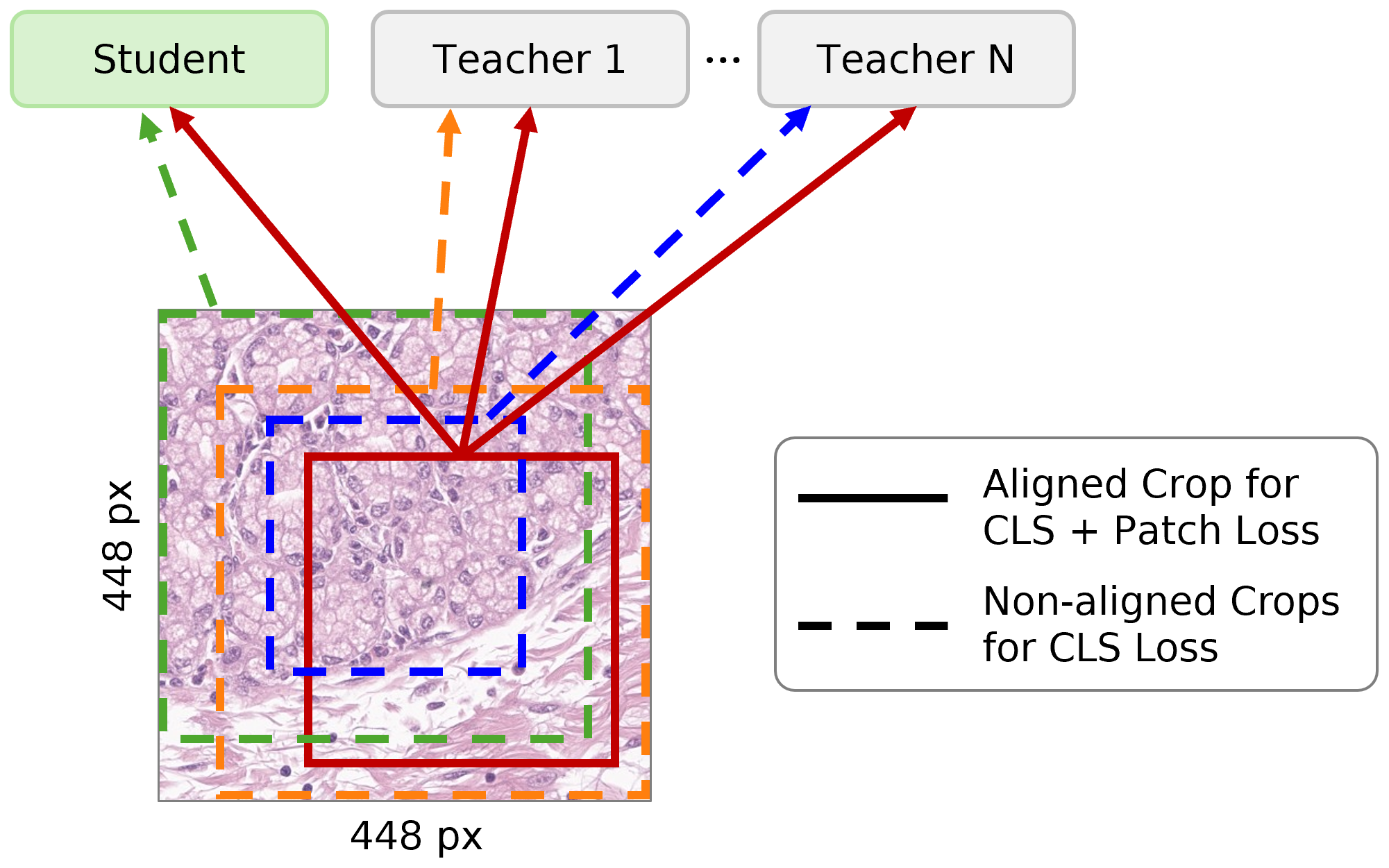}
   \caption{\textbf{Robustness-Enhancing Cropping Strategy.} A first random crop (red) is obtained from the input image. This \textit{aligned} crop is individually color augmented for the student and all teachers. Both CLS and patch loss are computed for this aligned crop. Additionally, each model receives a second random individual crop (dashed lines) which aims to improve magnification invariance. We only compute CLS loss on this \textit{non-aligned} crop.}
   \label{fig:cropping_strategy}
\end{figure}

\subsubsection{Data Augmentation.} Since the dataset originates from a single institution and was digitized using a single scanner, we apply extensive augmentations to enhance its diversity, including color augmentation, random flipping and Gaussian blur. For color augmentation, we employ H\&E-specific stain perturbations from the Albumentations library~\cite{info11020125} that decomposes an image into the hematoxylin and eosin channels before perturbing their intensities to simulate more realistic stain variations.
Additionally, we incorporate a novel cropping-based augmentation dedicated to our multi-teacher framework as explained in \cref{section:cropping} and \cref{fig:cropping_strategy}. 

\subsection{Pathryoshka Distillation Framework}

 Pathryoshka aims to obtain a compressed student model  by  distillation from multiple  teachers and training structured data representations. 
To this end, we introduce a novel cropping strategy to enhance robustness, employ multi-teacher nested embeddings, and design adjusted distillation heads suited to this architecture. The training objective is further supported by a dedicated loss formulation that aligns the student with the structured teacher signals. These  components of our framework are detailed in the following sections.

\subsubsection{Robustness-Enhancing Cropping Strategy.}
\label{section:cropping}
We generate multiple crops per tile to train the model (\cref{fig:cropping_strategy}). Each crop covers between 25\% and 100\% of the tile area with an aspect ratio sampled from $[0.9, 1.1]$, followed by random horizontal and vertical flips. The resulting region is resized to 224×224 px.
Formally, let $x$ denote an image tile from the dataset, $a_{\text{spatial}}$ the random cropping and flipping augmentation, and $a_{\text{visual}}$ a random H\&E stain and blur augmentation. The \textit{aligned crop} $x_c$ is defined as:
\begin{equation}
    x_c=a_{\text{spatial}}(x)
  \label{eq:aligned_crop}
\end{equation}

The student model and all teacher models receive a color augmented version of that aligned crop. Additionally, the student model and all teacher models receive a individual random crop called \textit{non-aligned crop} that is color augmented as well.
Those crops cover different areas of the original tile, effectively simulating variations in magnification and thereby improving robustness.
Formally, let $\mathcal{M} = {f_s, f_{t_1}, \dots, f_{t_N}}$ denote the set of student and teachers. Each model $f_m \in \mathcal{M}$ produces two outputs:
\begin{equation}
z_m^{\text{a}} = f_m(a_{\text{visual}}(x_c)), \
z_m^{\text{g}} = f_m(a_{\text{visual}}(a_{\text{spatial}}(x))).
  \label{eq:embeddings}
\end{equation}
Here, $z_m^{\text{a}}$ is the output for the aligned crop and $z_m^{\text{g}}$ the output for the non-aligned crop. Each output $z_m$ comprises the CLS token and patch embeddings of the respective model. For the aligned crop we compute a loss for both the CLS token and the patch tokens. For the non-aligned crop, we only compute a loss for the CLS token that doesn't require strict spatial correspondence. \\

Note, that \textit{DINOv2 distillation}~\cite{oquab2023dinov2} would require the teachers to forward the CLS token and patch tokens through the DINO and iBOT head, respectively, to obtain the high-dimensional teacher \textit{prototype scores}. However, the weights for the DINO and iBOT head are often not publicly available making DINOv2 distillation unfeasible.

\subsubsection{Nested Embeddings.}

We incorporate nested embeddings~\cite{kusupati2022matryoshka}, which allow the distilled model to generate representations that adapt to varying computational budgets and downstream requirements.

Given an input $x$, the distilled model produces an embedding $z \in \mathbb{R}^d$. We aim to learn embeddings where any prefix $z_{1:m}$, with $m \leq d$, forms a semantically meaningful representation. Let $M$ denote the set of target embedding sizes. For each $m \in M$, we enforce representational consistency through a loss-based constraint. Following ~\cite{venkataramanan2025franca}, we adopt a geometrically decreasing series of nested embedding sizes $M = \{768, 384, 192, 96, 48\}$, obtained by halving the dimension for each level. At training time, we compute distillation losses at all nesting levels, ensuring that truncated representations $z_{1:m}$ remain semantically aligned with teacher features.

\subsubsection{Distillation Heads.}

We employ three-layer MLPs to project the outputs of the distilled student to the embedding dimension of the respective teacher. The size of the first MLP layer matches the student embedding and the last layer the embedding size of the teacher model. We set the intermediate layer to the same dimensionality as the teacher embeddings to avoid creating an additional bottleneck. This design ensures that the only compression in the network occurs between the teacher and student embedding spaces.

Two sets of projection heads are used per teacher, one being for the CLS token, the other for the patch tokens encoding spatial information.
Each set of projection heads consists of $|M|$ MLPs, where $|M|$ denotes the number of nesting levels. Each MLP is responsible for one nesting level of dimension $m$ and receives the first $m$ features of the embedding.

\subsection{Loss Formulation} \label{loss_formulation}

The training targets of our student model are the outputs of multiple teacher models. Since all teachers are vision transformers, each produces a global CLS token, which captures image-level information, and a set of patch tokens that encode spatial features.

Let $x$ denote the augmented input image. The student network, parameterized by $\Theta$, outputs a sequence of tokens:
\begin{equation}
  \mathbf{z} = f(x | \Theta) = [z_s, z_p^{(1)}, z_p^{(2)}, \ldots, z_p^{(N)}],
  \label{eq:outputs_student}
\end{equation}
where $z_s \in \mathbb{R}^d$ is the global CLS token and $z_p^{(i)} \in \mathbb{R}^d$ are the $N$ patch tokens, each representing a spatial region of the input.

Similarly, the output for teacher $t$ with parameters $\Theta_t$ is denoted as
\begin{equation}
  \mathbf{z^t} = f_t(x | \Theta_t) = [z_s^{(t)}, z_p^{(t,1)}, z_p^{(t,2)}, \ldots, z_p^{(t,N)}],
  \label{eq:outputs_teacher}
\end{equation}
where $z_s^{(t)}$, $ z_p^{(t,i)} \in \mathbb{R}^{d_t}$.

For every teacher $t$, we attach two families of distillation heads:
\begin{equation}
    g_s^{(t,m)}: \mathbb{R}^m \rightarrow \mathbb{R}^{d_t}, \quad
    g_p^{(t,m)}: \mathbb{R}^m \rightarrow \mathbb{R}^{d_t},
\end{equation}
where $m \in M$ denotes the nesting dimension (i.e., the subset of the first m features) and $M$ is the set of considered nesting levels. Each $g_s^{(t,m)}$ projects the first $m$ dimensions of the student’s CLS token to the teacher’s embedding space, while $g_p^{(t,m)}$ performs an analogous projection for each patch token.

The projected features are defined as:
\begin{equation}
  \hat{z}_s^{(t,m)} = g_s^{(t,m)}(z_{s,1:m}),
  \label{eq:projection_summary}
\end{equation}
\begin{equation}
  \hat{z}_p^{(t,i,m)} = g_p^{(t,m)}(z_{p,1:m}^{(i)}),
  \label{eq:projection_patch}
\end{equation}
where $z_{s,1:m}$ and $z_{p,1:m}^{(i)}$ denote the first $m$ dimensions of the corresponding embeddings.

The loss for the CLS token is computed as:
\begin{equation}
  \mathcal{L}_{\text{cls}} = \sum_t \sum_m \mathcal{L}_{\text{cos}}(\hat{z}_s^{(t,m)}, z_s^{(t)}),
  \label{eq:cls_loss}
\end{equation}
where $\mathcal{L}_{\text{cos}}$ is the cosine similarity loss and $z_s^{(t)}$ denotes the teacher’s CLS token.

Before computing the patch-level loss, we standardize each teacher’s patch embeddings to account for differences in embedding magnitude, following~\cite{ranzinger2024phi}:
\begin{equation}
  \tilde{z}_p^{(t,i)} = \frac{z_p^{(t,i)} - \mu_c^t}{\sigma_c^t},
  \label{eq:standardization}
\end{equation}
where $\mu_c^t$ and $\sigma_c^t$ are the mean and standard deviation computed over channel $c$ of teacher $t$ for the current batch.

Finally, the patch-level distillation loss is defined as:
\begin{equation}
  \mathcal{L}_{\text{patch}} = \sum_t \sum_i \sum_m \mathcal{L}_{\text{MSE}}(\hat{z}_p^{(t,i,m)}, \tilde{z}_p^{(t,i)}),
  \label{eq:patch_loss}
\end{equation}
where $\mathcal{L}_{\text{MSE}}$ denotes the mean squared error loss, utilized here in accordance with established distillation paradigms~\cite{ranzinger2024radio}.

The loss formulation would allow assigning a weight $\lambda_t$ to each teacher. Considering the large exploration space and the significant training cost, we chose to give each teacher equal weight. This ensures that the student model learns a balanced representation derived from the collective expertise of the ensemble.

\subsection{Teacher Selection}
\label{sec:teacher_selection}
Today, a plethora of pathology FMs exist~\cite{wang2022transformer, vorontsov2024foundation, chen2024uni, nechaev2024hibou, luVisuallanguageFoundationModel2024, zimmermann2024virchow2, hoptimus0, filiotScalingSelfSupervisedLearning2023, filiotPhikonv2LargePublic2024, xu2024whole, huang2023visual, zhang2025multimodal, karasikov2025training}. We selected a representative ensemble of the most recent versions of high-capacity teachers: Virchow2~\cite{zimmermann2024virchow2}, UNI2-h~\cite{chen2024uni} and H-optimus-1~\cite{hoptimus1}. These models were chosen for their state-of-the-art performance across diverse pathological benchmarks \cite{neidlinger2025benchmarking}, with each exceeding 600 million parameters. 

All of them are trained by distinct institutions with proprietary data increasing their heterogeneity compared to models trained on overlapping public datasets. Their complementarity is confirmed in the evaluation in \cref{tab:benchmark}. There is no single teacher dominating in all benchmark datasets.

To quantify the representational overlap, we computed the centered linear kernel alignment (CKA)~\cite{kornblith2019similarity} between the teacher models on a subset of 10,000 images of our training dataset. CKA measures the similarity between representations while being invariant to linear transformations and ranges from 0 (orthogonal) to 1 (identical). We observed pairwise CKA scores between 0.64 and 0.71 for our pathology-specific teacher models suggesting that their representations are domain-aligned but can contribute distinctive views. In comparison, a ViT-L model pre-trained with DINOv2 on the LVD142M natural image dataset~\cite{oquab2023dinov2} and our selected teachers have pairwise CKA scores between 0.35 and 0.40. To obtain a baseline for model-to-model consistency, we measured CKA of two models with different sizes (ViT-L and ViT-g) pre-trained on the same dataset (LVD142M) yielding a high similarity score of 0.88 for our subset.

\section{Experiments and Results}

We evaluate \modelname{} in two sizes: \modelname-B (86M parameters) and \modelname-S (22M), and compare them to their teacher models, to the state-of-the-art single-teacher distilled pathology FM, H0-mini, and to the multi-teacher model GPFM. Because H0-mini and GPFM are distilled models, they are serving as baselines for single- and multi-teacher distillation. Furthermore, we evaluate how the different nesting levels behave compared to FMs without such an ordering in their embeddings.

\subsection{Classification Benchmark Results}

\begin{table}[tb]
\caption{Benchmark on public datasets for our models in comparison to the baselines H0-mini and GPFM, and the teacher models Virchow2, UNI2-h and H-optimus-1. We report average multiclass accuracy for multi-class classification, and binary balanced accuracy for binary classification tasks over five runs as reported by the \textit{eva} framework~\cite{kaiko.ai2024eva}. \textsuperscript{\textdagger}~indicates that the dataset was part of the training data of this model, therefore generalization may be overestimated. Highlighted are \textbf{best} and \underline{second best} models.}
\centering

\fontsize{7pt}{8.5pt}\selectfont
\begin{tabular}{lcccccccc}
\toprule
 & \multicolumn{3}{c}{\textbf{\textls[200]{Teacher}}} & \multicolumn{2}{c}{} & \multicolumn{2}{c}{\textbf{\textls[200]{Ours}}} \\ 
\cmidrule(lr){2-4}
\cmidrule(lr){7-8}
\textbf{Dataset} & Virchow2 & UNI2-h & H-optimus-1 & H0-mini & GPFM & \textbf{Pathryoshka-S} & \textbf{Pathryoshka-B} \\
\midrule
BACH & 88.0 \res{0.5} & \underline{90.7} \res{1.1} & 78.1 \res{0.8} & 79.2 \res{0.5} & 83.0\textsuperscript{\textdagger}\res{0.4} & 85.8 \res{0.5} & \textbf{90.8} \res{0.3} \\
BRACS & 62.2 \res{0.7} & \textbf{66.1} \res{0.9} & 61.9 \res{0.5} & 60.6 \res{0.6} & 62.8\textsuperscript{\textdagger}\res{0.5} & 61.7 \res{1.0} & \underline{65.3} \res{0.2} \\
BreakHis & 81.9 \res{0.6} & \underline{85.9} \res{0.3} & 76.0 \res{1.2} & 74.7 \res{0.9} & 76.0\textsuperscript{\textdagger}\res{0.5} & 76.2 \res{0.4} & \textbf{87.1} \res{0.7} \\
CRC & \underline{96.6} \res{0.2} & 96.5 \res{0.2} & 95.5 \res{0.0} & 96.1 \res{0.2} & 95.2\textsuperscript{\textdagger}\res{0.1} & 96.0 \res{0.0} & \textbf{96.7} \res{0.1} \\
Gleason & \underline{77.9} \res{0.6} & 77.4 \res{0.3} & 78.5 \res{0.5} & \underline{77.9} \res{0.5} & 76.6 \res{0.6} & \textbf{78.7} \res{0.8} & \textbf{78.7} \res{0.6} \\
MHIST & \textbf{86.3} \res{0.1} & 82.6 \res{0.3} & 83.7 \res{0.3} & 79.0 \res{0.2} & 81.4 \res{0.1} & 82.2 \res{0.4} & \underline{84.0} \res{0.2} \\
Patch Cam. & 93.8 \res{0.1} & \textbf{95.1} \res{0.1} & 94.2 \res{0.1} & 94.2 \res{0.1} & \underline{94.4} \res{0.0} & 92.1 \res{0.1} & 94.1 \res{0.1} \\
CAM. 16 & 93.1 \res{1.4} & 93.8 \res{1.3} & 92.1 \res{1.3} & 93.1 \res{1.7} & \underline{94.9}\textsuperscript{\textdagger}\res{2.6} & 93.5 \res{1.3} & \textbf{95.0} \res{1.0} \\
PANDA & 75.8 \res{1.0} & 76.8 \res{1.1} & 77.2 \res{1.4} & \underline{78.1} \res{0.4} & \textbf{79.1}\textsuperscript{\textdagger}\res{0.9} & 76.2 \res{0.9} & 77.0 \res{0.8} \\
UniToPatho & \textbf{57.5} \res{0.3} & 54.0 \res{0.3} & 48.7 \res{0.6} & 49.2 \res{0.3} & 49.7\textsuperscript{\textdagger}\res{0.1} & 52.7 \res{0.4} & \underline{54.9} \res{0.1} \\
\midrule
\textbf{Median} & 84.1 & \underline{84.3} & 78.3 & 78.6 & 80.3 & 80.5 & \textbf{85.6} \\
\bottomrule
\end{tabular}
\label{tab:benchmark}
\end{table}

\begin{table}[tb]
\caption{Comparison of computational efficiency across our models (Path.-S/B), teacher models, and baselines including parameter count, FLOPs, and throughput measured on an RTX 3090 GPU.}
\centering
\fontsize{8pt}{9pt}\selectfont
\begin{tabular}{lccccccc}
\toprule
& Virchow2 & UNI2-h & H-optimus-1 & H0-mini & GPFM & \textbf{Path.-S} & \textbf{Path.-B} \\
\midrule
\textbf{Params.}  [M] $(\downarrow)$ & 631 & 681 & 1,135 & 86 & 303 & \textbf{22} & 86 \\
\textbf{FLOPs} [G] $(\downarrow)$ & 329.1 & 360.7 & 591.8 & 44.6 & 155.6 & \textbf{11.1} & 44.6 \\
\textbf{Throughput} [img/s] $(\uparrow)$ & 139 \res{2} & 136 \res{1} & 82 \res{1} & 842 \res{4} & 278 \res{1} & \textbf{2338} \res{14} & 842 \res{4}\\
\bottomrule
\end{tabular}
\label{tab:performance}
\end{table}

We evaluate all models on ten commonly used public pathology benchmarks. Eight of those are patch-level tasks (BACH~\cite{aresta2019bach}, BRACS~\cite{brancati2022bracs}, BreakHis~\cite{spanhol2015dataset}, CRC 100k~\cite{Kather2018Dataset}, Gleason Grading~\cite{arvaniti2018automated}, MHIST~\cite{wei2021petri}, Patch Camelyon~\cite{veeling2018rotation}, and UniToPatho~\cite{barbano2021unitopatho}), and two are slide-level tasks (CAMELYON16~\cite{bejnordi2017diagnostic} and PANDA~\cite{bulten2022artificial}). The selected benchmarks cover a wide range of magnifications levels from 0.25 microns per pixel (mpp) to 3.5 mpp. For all public benchmarks we used the \textit{eva}~\cite{kaiko.ai2024eva} benchmark framework. This framework uses a single linear layer for patch classification tasks and an ABMIL~\cite{ilse2018attention} head for slide-level classification tasks to classify the CLS embedding of the respective model. To ensure a fair comparison of our models with the baseline and the teachers, we run the benchmarks for all models using \textit{eva}'s default configuration. We report the average of five runs of the multiclass classification accuracy for multiclass tasks and binary balanced accuracy for binary classification tasks. The results are shown in Table \ref{tab:benchmark}.

Compared to its much larger teacher models,  \modelname-B delivers competitive performance across the benchmark datasets.
It exceeds the median accuracy of UNI2-h by 1.3 points, Virchow2 by 1.5 points, and H-optimus-1 by 7.3 points, despite requiring 86\%, 87\%, and 92\% fewer parameters, respectively.
Furthermore, it surpasses the distilled models by 7.0 (H0-mini) and by 5.3 (GPFM) points in median accuracy, while additionally providing nested embeddings. With a four-fold reduction in parameters compared to \modelname-B, \modelname-S shows a 5.1-point drop in median accuracy. However, its median balanced accuracy still exceeds that of H0-mini, H-optimus-1, and GPFM. The advantage of \modelname{} over the  top-performing teacher model UNI2-h, as well as H0-mini and GPFM baselines is statistically significant as determined using a one-sided Wilcoxon Signed-Rank Test~\cite{demvsar2006statistical} across all datasets with
 p-values of $p=0.042$, $p = 0.010$, and $p = 0.019$ respectively. Tests for teacher superiority over Pathryoshka-B yield consistently high $p$-values (UNI2-h: $0.967$, Virchow2: $0.935$, H-Optimus-1: $0.995$), confirming the teachers fail to significantly outperform our student.

\begin{table}[tb]
\caption{Random cropping augmentation matches or exceeds performance framework on 9 out of 10 benchmarks, showing distinct improvements in 7 instances.}
\label{tab:augmentations_ablation}
\centering
\fontsize{8pt}{9pt}\selectfont
\begin{tabular}{lccccccc}
\toprule
\multirow{2}{*}{\textbf{Dataset}} & Pathryoshka-B & Pathryoshka-B \\
& \textbf{crop} & \textbf{no crop} \\
\midrule
BACH & \textbf{90.8} \res{0.3} & 90.0 \res{0.4} \\
BRACS &  \textbf{65.3} \res{0.2}& 64.6 \res{0.7} \\
BreakHis & \textbf{87.1} \res{0.7}& 85.5 \res{0.6} \\
CRC & \textbf{96.7} \res{0.1} & 96.0 \res{0.1} \\
Gleason & \textbf{78.7} \res{0.6}& 77.7 \res{0.3} \\
MHIST & 84.0 \res{0.2} & \textbf{85.1} \res{0.2} \\
Patch Cam. & \textbf{94.1} \res{0.1} & 93.6 \res{0.0} \\
CAM. 16 & \textbf{95.0} \res{1.0} & 91.6 \res{1.5} \\
PANDA & \textbf{77.0} \res{0.8} & \textbf{77.0} \res{1.2} \\
UniToPatho & \textbf{54.9} \res{0.1} & \textbf{54.9} \res{0.2} \\
\midrule
\textbf{Average} & \textbf{82.4} & 81.6\\
\textbf{Median} & \textbf{85.6} & 85.3 \\
\bottomrule
\end{tabular}
\end{table}

An ablation study of the random-cropping strategy introduced in Section \ref{sec:method} demonstrates its effectiveness: incorporating random cropping improves the average accuracy of Pathryoshka-B by 0.8 points and median by 0.2 across the public benchmarks as shown in Table~\ref{tab:augmentations_ablation}. Another ablation of the number of teachers  demonstrated that a single teacher variant inherited its teacher’s weaknesses and performed  worse in median balanced accuracy than our multi-teacher distilled model (Supplementary Table S6).

Table~\ref{tab:performance} compares parameter counts and inference efficiency of the evaluated models, reporting  FLOPs and throughput (images per second) measured on a consumer-grade RTX 3090 GPU. Throughput is computed in FP16 precision with a batch size of 32, and we report the mean and standard deviation over 500 batches. Our best-performing model, Pathryoshka-B, uses 92\% fewer parameters and achieves roughly 11$\times$ higher throughput compared to its largest teacher, H-optimus-1. In whole slide image analysis, where thousands of image tiles must be encoded for a single prediction, this efficiency translates into substantial reductions in total processing time.

\subsection{Evaluation of Nested Embeddings}
\label{sec:eval_nested_embeddings}

\begin{figure}[tb]
  \centering
   \includegraphics[width=1.0\linewidth]{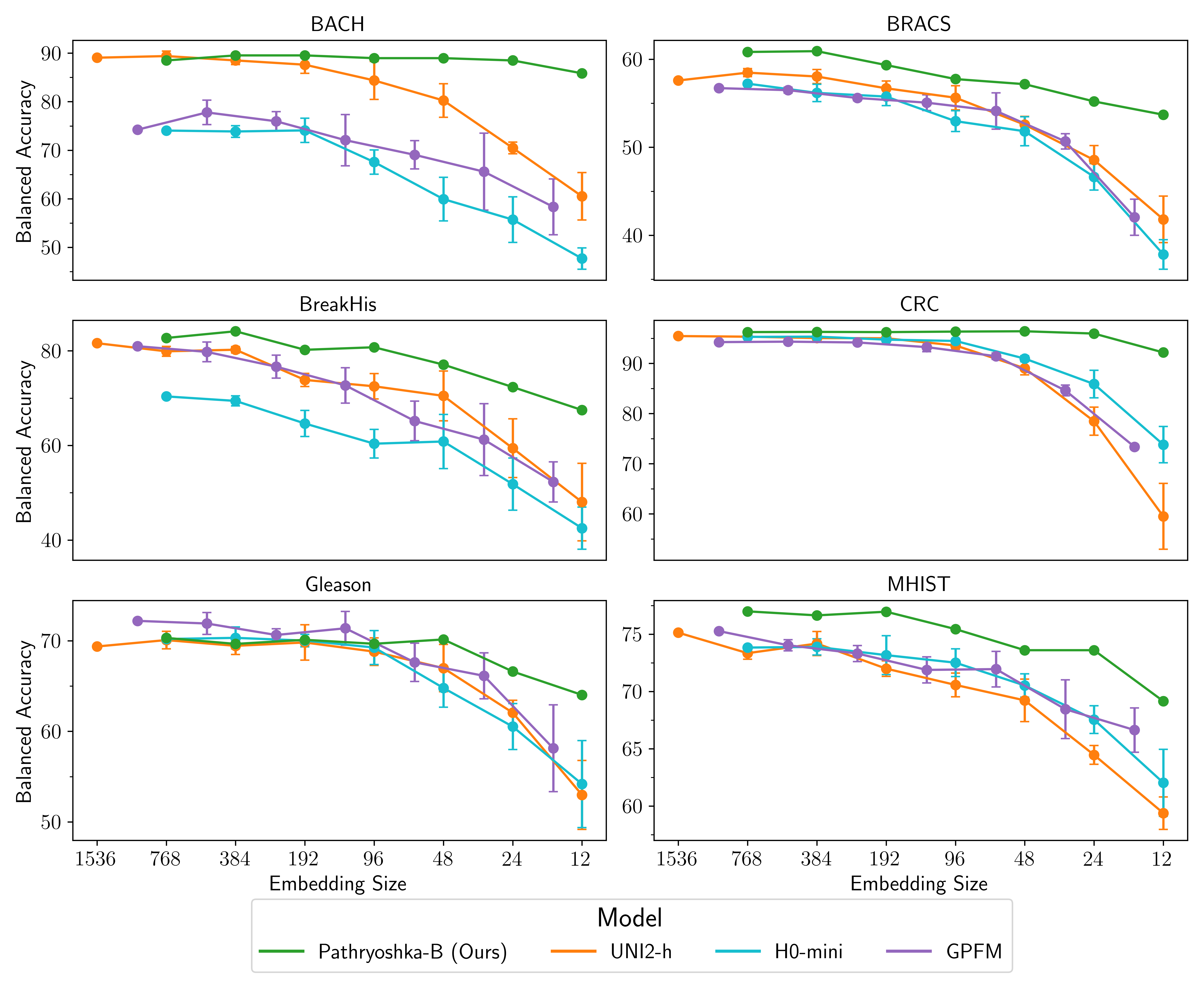}
   \caption{Balanced k-NN classification accuracy between  \modelname-B, the best-performing teacher model UNI2-h, and the baselines H0-mini and GPFM, across six patch-based multiclass classification benchmarks. Each model is evaluated using successively halved subsets of its original embedding, down to a minimum embedding size of 12. Error bars indicate the standard deviation across five random feature sampling runs for the baseline models. Benefiting from its nested embedding structure, accuracy for \modelname-B degrades gracefully even at severely reduced dimensionalities.}
   \label{fig:knn_classification_accuracy}
\end{figure}

\subsubsection{k-NN-Classification.}

In ~\cref{fig:knn_classification_accuracy}, we benchmark the quality of the learned representations by progressively reducing the embedding size. We evaluate k-NN classification ($k=10$, cosine similarity) on six multiclass patch classification datasets. \modelname-B is compared against the best-performing teacher model, UNI2-h, and the two baselines H0-mini and GPFM. Note that the full embedding dimension varies across models: UNI2-h uses 1536, GPFM uses 1024, while both Pathryoshka-B and H0-mini use 768.

Due to its nested embedding design, the features of our model are inherently ordered by relevance. To assess this property, we evaluate performance using only the first $\frac{\text{dim}}{n}$ features of our model for $n \in \{1, 2, 4, 8, 16, 32, 64\}$. To ensure a fair comparison, we match these dimensionalities for the other models by randomly sampling $\frac{dim}{n}$ features, reporting the average over five runs. For UNI2-h, we additionally evaluate $n=128$ so that its lowest feature count matches ours. Detailed results are provided in the Supplementary Material in Table S4.

Interestingly, performance remains relatively stable for all models up to $\frac{\text{dim}}{4}$, suggesting a degree of redundancy in the embeddings of UNI2-h, H0-mini and GPFM. Beyond this threshold, UNI2-h, H0-mini and GPFM exhibit a significant drop in accuracy. In contrast, Pathryoshka-B degrades much more gracefully. The average relative performance decrease from the full embedding down to $\frac{dim}{64}$ across the six datasets is only 9.4\% for our model, compared to 17.7\% (UNI2-h), 28.4\% (H0-mini) and 22.64\% (GPFM). At an embedding size of only 12 it is outperforming the other models by a large margin. Notably, even though Pathryoshka-B is  explicitly trained only for nested sizes down to $\frac{\text{dim}}{16}$, further truncated embeddings still maintain competitive accuracy.

\begin{figure}[tb]
    \centering
    \includegraphics[width=0.95\linewidth]{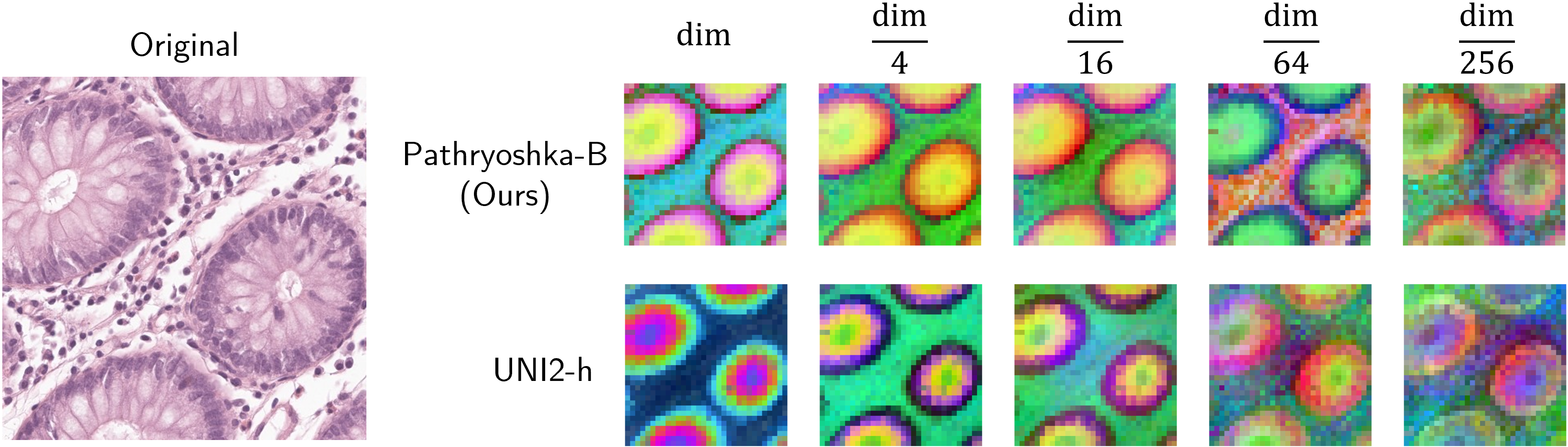}
    \caption{
        Visualization of the semantic degradation of patch features when truncating the embeddings of \modelname-B and UNI2-h, by mapping the first three PCA components to the RGB color space. For the given image of intestinal glands from our training dataset, our model is able to sharpely delineate glands from connective tissue even with a dimensionality of three. For UNI2-h, which does not have a nested embedding structure, the glands start to blend in with surrounding tissue below $\frac{dim}{64}$.
    }
    \label{fig:embedding_visual}
\end{figure}

This behavior is further validated by the visualization of patch embeddings in Figure \ref{fig:embedding_visual}. Despite significant truncation, \modelname-B preserves semantic consistency in its patch tokens. To demonstrate this, we projected the first $\frac{dim}{n}$ dimensions ($n\in\{1, 4, 16, 64, 256\}$) into RGB using PCA. When only a fraction of the original dimensions remain, UNI2-h exhibits inconsistent representations for similar histological structures while \modelname-B maintains morphological consistency.

\subsubsection{Patch Retrieval.}

\begin{table}[tb]
\caption{Patch-retrieval experiment results on the CCRCC dataset. We report the average Recall@$5$ over all queries. Results for $K = 1$ and $K = 10$ are provided in the Supplementary Material Tab. S1. \textsuperscript{\textdagger}~indicates that CCRCC is contained in the training data. Pathryoshka-B exhibits minimal degradation when the feature dimensionality is reduced compared to the larger teachers and baseline models.}
\centering
\fontsize{8pt}{9pt}\selectfont
\begin{tabular}{lccccccc}
\toprule
\textbf{Model} & dim  & $\frac{\text{dim}}{2}$ & $\frac{\text{dim}}{4}$  & $\frac{\text{dim}}{8}$ & $\frac{\text{dim}}{16}$  & $\frac{\text{dim}}{32}$& $\frac{\text{dim}}{64}$\\
\midrule
Virchow2 & \underline{97.2}  &\underline{97.2}  &\underline{97.1} &\underline{97.1} &\textbf{97.0}&\textbf{96.6} & \underline{94.8}\\
UNI2-h & \underline{97.2} & 97.1 & 97.0 & 97.0 & 96.7 & 95.6 & 92.4\\
H-optimus-1 & 97.1 & 97.1  & 97.0  & 97.0 &96.6 &\underline{95.7} &92.7\\
\midrule
H0-mini\textsuperscript{\textdagger} & \textbf{97.4} & \textbf{97.3}  & \textbf{97.2} & 97.0 & 96.6 & 95.5 & 92.6 \\
GPFM\textsuperscript{\textdagger} & 97.3 & \textbf{97.3}  & \textbf{97.2} & \textbf{97.2} & 96.9 & 95.7 & 92.0 \\
\textbf{Pathryoshka-B} & 97.1 & 97.1 & 97.0 & 97.0 & \underline{96.9} & \textbf{96.6} & \textbf{95.2} \\

\bottomrule
\end{tabular}
\label{tab:img_ret}
\end{table}

We further evaluate the quality of nested embeddings via a patch retrieval task. For each query patch, the top-$k$ nearest patches are retrieved based on cosine similarity, and accuracy is measured by label agreement with the query. The experiment is conducted on the CCRCC dataset~\cite{brummer2023computational}. Table~\ref{tab:img_ret} shows Pathryoshka-B matches the retrieval performance of large-scale teacher models while surpassing all baselines at compressed embedding sizes of $\frac{dim}{32}$ and $\frac{dim}{64}$.

\begin{table}[ht]
\caption{Evaluation of nested embeddings of the patch tokens on segmentation tasks using the ConSep and MoNuSAC dataset. We compare the performance of \modelname{}-B with full and reduced embedding size of dim/16 against GPFM and H0-mini using the average Dice score of five runs.}
\centering
\fontsize{8pt}{9pt}\selectfont
\begin{tabular}{lcccc}
\toprule
& \multicolumn{2}{c}{\textbf{ConSep}} & \multicolumn{2}{c}{\textbf{MoNuSAC}} \\
\cmidrule(lr){2-3} \cmidrule(lr){4-5}
\textbf{Model} & full dim & dim/16 & full dim & dim/16 \\
\midrule
Pathryoshka-B & \textbf{63.6} \res{0.3} & \textbf{62.6} \res{0.2} & \underline{63.8} \res{0.7} & \textbf{61.3} \res{0.8} \\
GPFM          & 47.8 \res{0.2} & 47.6 \res{0.4} & 48.3 \res{0.4} & 47.4 \res{0.8} \\
H0-mini       & \underline{63.1} \res{0.4} & \underline{54.8} \res{0.1} & \textbf{64.0} \res{1.0} & \underline{55.4} \res{1.0} \\
\bottomrule
\end{tabular}

\label{tab:segmentation}
\end{table}

\subsubsection{Segmentation.} 
Finally, we performed quantitative verification of \modelname's nested patch tokens using the \textit{eva} evaluation framework and reporting the average Dice score of five independent runs for the ConSep~\cite{graham2019hover} and MoNuSAC~\cite{verma2021monusac2020} datasets. In this task, we compared to the two distilled pathology FM baselines, H0-mini and GPFM. 
The segmentation masks are produced by forwarding the patch tokens and the raw input image to a task-specific head featuring a lightweight convolutional decoder. The resulting performance is detailed in \cref{tab:segmentation}. At a heavily reduced embedding size of $\frac{\text{dim}}{16}$, our nested model maintains robust accuracy with Dice score drops restricted to 1.6\% (ConSep) and 3.9\% (MoNuSAC), while significantly outperforming both distilled baselines.

 \section{Discussion and Conclusion}
In this work, we introduced \modelname{}, a pathology foundation model that integrates multi-teacher distillation with nested embeddings. By employing a multi-teacher strategy with 
cropping that fosters robustness and distillation-adapted augmentation, \modelname{} integrates complementary knowledge from multiple experts, mitigating the individual biases inherent in single-teacher distillation to produce a more robust and generalizable student.  Furthermore, the implementation of nested embeddings provides a capability to reduce embedding size based on downstream task requirements enabling a trade-off between accuracy and computational cost. It allows for severe truncation  of the embedding with only minimal degradation in accuracy.
While validated on digital pathology, our approach is domain-agnostic and applicable to other fields.

From a deployment perspective, the model is already well suited for clinical settings: \modelname{} delivers a lightweight architecture and fast inference while maintaining parity with state-of-the-art large-scale FMs and consistently outperforming state-of-the-art single-teacher and multi-teacher distillation pathology FMs. While our evaluation already masters well-established classification and representation benchmarks capturing many practical needs in pathology, future work includes more extensive evaluation of clinical utility. 

Our approach challenges the conventional scaling paradigm, demonstrating that performance does not strictly depend on increasing model parameters or dataset volume.   These findings suggest that 
current large-scale FMs underutilize their capacity, with high-dimensional embeddings containing significant redundancy. These results underscore the need for more efficient pre-training strategies over raw scaling, a direction we hope will spark further research in the field.

\section*{Acknowledgements}
We acknowledge support by the Munich Data Science Institute (MDSI) at Technical University of Munich (TUM) via the MDSI Doctoral Fellowship program, and the BMFTR-funded SATURN3 project (01KD2206C).

%
%
\bibliographystyle{splncs04}
\bibliography{main}

\end{document}